\documentclass[runningheads]{llncs}
\usepackage{graphicx}

\usepackage[ruled,vlined]{algorithm2e}
\usepackage{multirow}
\usepackage{tikz}
\usepackage{comment}
\usepackage{amsmath,amssymb} 
\usepackage{color}
\usepackage{orcidlink}

\usepackage[accsupp]{axessibility}  

\begin{document}
\pagestyle{headings}
\mainmatter
\def\ECCVSubNumber{135}  

\title{PAI3D: Painting Adaptive Instance-Prior for 3D Object Detection} 

\titlerunning{PAI3D: Painting Adaptive Instance-Prior for 3D Object Detection}
%
\author{
  Hao Liu \and Zhuoran Xu \and Dan Wang \and Baofeng Zhang \and Guan Wang \and Bo Dong \and Xin Wen \and Xinyu Xu
}
\authorrunning{JDXAD Perception Team}
%
\institute{Autonomous Driving Division of X Research Department, JD Logistics\\
\email{\{liuhao163, xuzhuoran, wangdan257, zhangbaofeng13, wangguan151, dongbo14, wenxin16, xinyu.xu\}@jd.com}}

\maketitle

\begin{abstract}
  3D object detection is a critical task in autonomous driving. Recently multi-modal fusion-based 3D object detection methods, which combine the complementary advantages of LiDAR and camera, have shown great performance improvements over mono-modal methods. However, so far, no methods have attempted to utilize the instance-level contextual image semantics to guide the 3D object detection. In this paper, we propose a simple and effective Painting Adaptive Instance-prior for 3D object detection (PAI3D) to fuse instance-level image semantics flexibly with point cloud features. PAI3D is a multi-modal sequential instance-level fusion framework. It first extracts instance-level semantic information from images, the extracted information, including objects categorical label, point-to-object membership and object position, are then used to augment each LiDAR point in the subsequent 3D detection network to guide and improve detection performance. PAI3D outperforms the state-of-the-art with a large margin on the nuScenes dataset, achieving 71.4 in mAP and 74.2 in NDS on the test split. Our comprehensive experiments show that instance-level image semantics contribute the most to the performance gain, and PAI3D works well with any good-quality instance segmentation models and any modern point cloud 3D encoders, making it a strong candidate for deployment on autonomous vehicles.
  \end{abstract}

  \footnotetext[1]{The authors share equal contributions.}

  \section{Introduction}
  \begin{figure}[t]
    \centering
    \includegraphics[height=4.8 cm]{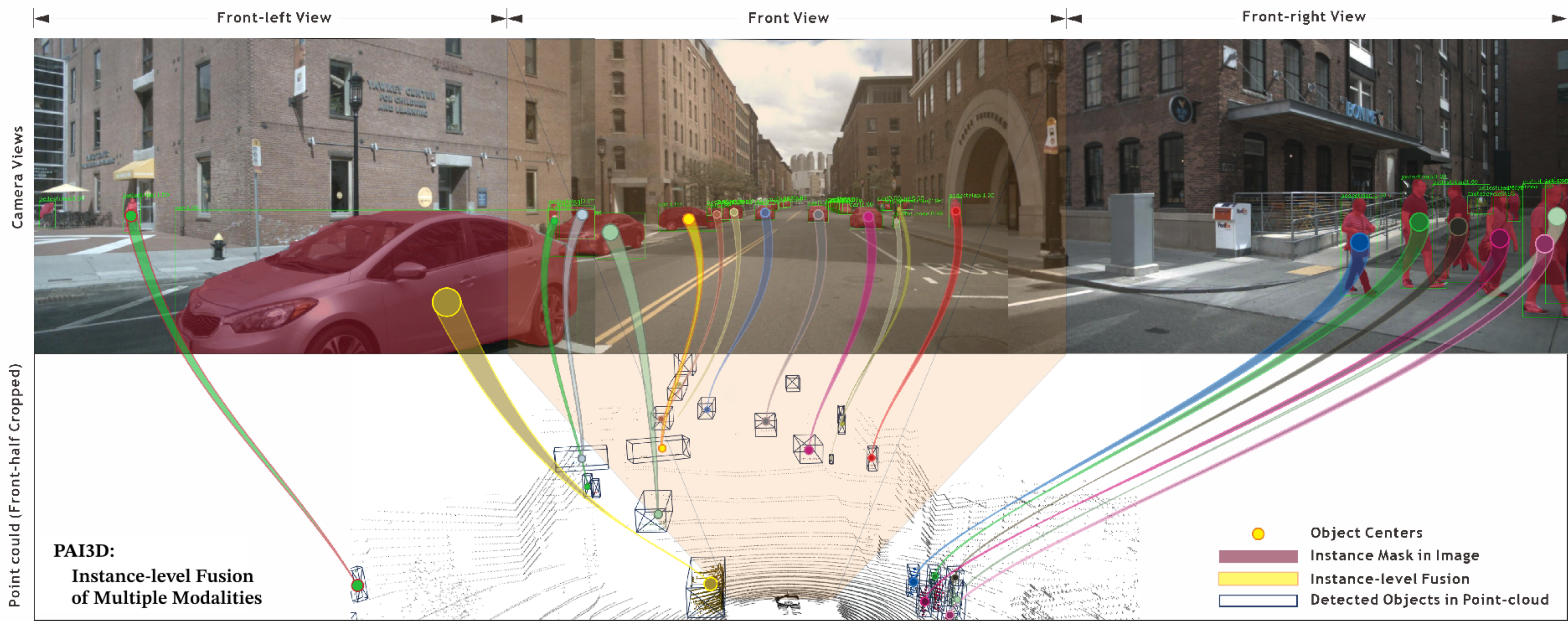}
    \caption{PAI3D fuses point cloud and images at instance-level for 3D object detection. We utilize image instance segmentation to obtain 3D instance priors, including instance-level semantic label and 3D instance center, to augment the raw LiDAR points. We select one frame from the nuScenes dataset and show the front half point cloud and the three frontal image views. Best viewed in color.
    }
    \label{fig:PAI3D_illus}
  \end{figure}

  3D object detection is critical for autonomous driving. LiDAR, camera and Radar are the typical sensors equipped on autonomous driving cars to facilitate 3D object detection. Up to date point cloud alone based 3D object detection methods \cite{pointpillars,pixor,RSN,voxelnet,yan2018second,centerpoint} achieve significantly better performance than camera alone monocular 3D object detection methods \cite{smoke,Monopair,Fcos3d} in almost all large-scale Autonomous Driving Dataset such as nuScenes \cite{nuscenes} and Waymo Open Dataset \cite{waymo}, making LiDAR the primary, first-choice sensor due to its accurate and direct depth sensing ability. However, point cloud data may be insufficient in some scenarios, e.g., where objects are far away with very few points reflected, or objects are nearby but are very small or thin. Recently more and more works employ both point cloud and images for 3D object detection \cite{MV3D,AVOD,3D-CVF,yang2018ipod,PointPainting,ding20201st,FusionPainting,mvp,park2021multi}, aiming to utilize the complementary characteristics of 3D and 2D sensors to achieve multi-modal fusion, and these methods have demonstrated great performance boost over mono-modal approaches.

  Existing multi-modal fusion methods either predict semantic labels from images, which are used as semantic priors to indicate foreground points in 3D point cloud \cite{PointPainting,FusionPainting}, or incorporate implicit pixel features learned from image encoder into the 3D detection backbone \cite{AVOD,MV3D}. Since the objective of 3D object detection is to identify each individual object instance, most of these methods, however, fail to reveal the instance-level semantics from 2D images, and hence hardly provide clear guidance to distinguish instance-level features in 3D, especially for the objects in the same category. Therefore, it is necessary to go beyond the existing category-level fusion and explore deep into the instance-level fusion for more explicit detection guidance.

  Along the line of extracting instance-level information to improve 3D object detection, previous studies \cite{centerpoint,centernet} have shown that representing and detecting objects as points is effective. These methods focus on using centers to facilitate ground-truth matching during training for only mono-modality, either point cloud or images, and the pros and cons of utilizing instance center from an additional modality and sharing it throughout the detection pipeline hasn't been studied yet. Our intuition is that extracting instance centers from images as instance priors and then feeding these instance priors into a 3D object detection network for a deep fusion with the raw LiDAR points will be highly beneficial for the 3D object detection task. However, a naive, straight-forward implementation of such idea may yield degraded performance due to erroneous semantic labels caused by point-to-pixel projection errors. Such errors can occur easily in an autonomous driving system equipped with multi-modal sensors.

  To address the above challenges, in this paper, we propose a novel sequential multi-modal fusion method named PAI3D, to utilize instance priors extracted from image instance segmentation to achieve improved 3D object detection. The term “PAI” stands for Painting Adaptive Instance-prior from 2D images into 3D point cloud, where “Adaptive” refers to adaptively dealing with erroneous semantic labels caused by point-to-pixel projection errors. Fig. \ref{fig:PAI3D_illus} schematically illustrates the instance-level fusion between point cloud and images.

  As shown by the PAI3D architecture in Fig. \ref{fig:overall}, PAI3D includes three key learnable modules. Instance Painter predicts instance-level semantic labels from images, it then augments the raw point cloud with these instance-level semantic labels by point-to-pixel projection. Since the augmented point cloud sometimes contains erroneous instance labels caused by projection mismatch introduced by inaccurate sensor calibration, object occlusions or LiDAR-camera synchronization errors, we design Adaptive Projection Refiner to correct the erroneous instance labels and output a 3D center as instance prior for each augmented LiDAR point. Moreover, road scene objects can exhibit large-scale variations, ranging from long buses or trucks to small pedestrians. Although good performance can still be achieved by detecting objects of similar scale from a single feature map associated with one detection head \cite{CBGS}, it is generally difficult to accurately detect objects with large-scale variations from a single feature map associated with one detection head. To overcome this difficulty, we propose Fine-granular Detection Head with multiple feature maps of varying receptive fields to localize specific-scaled objects from specific-scaled feature maps, which greatly improves detection accuracy for objects with large-scale variations.

  To summarize, the main contributions of this paper include the following.

  (1) We propose a novel instance-level fusion method for 3D object detection, PAI3D, where instance-level semantics, including semantic label and 3D instance center, are utilized as priors to augment the raw LiDAR points, resulting in significant performance improvement.

  (2) Adaptive Projection Refiner is proposed to adaptively correct the erroneous instance-level semantics, which makes the proposed method more robust and tolerant to imperfect sensor calibration, objects occlusions and LiDAR-camera synchronization errors.

  (3) We design Fine-granular Detection Head of varying receptive fields to detect objects with varying scales. This module creates fine-granular match between corresponding object scale and receptive field in multi-resolution feature maps, enabling more accurate detection of objects with large-scale variations.

  We first evaluate PAI3D on the well-known large-scale nuScenes dataset \cite{nuscenes}. PAI3D outperforms the state-of-the-art (SOTA) with a large margin, achieving 71.4 mAP and 74.2 NDS on the test split (no external data used). We then present several ablation results to offer insights into the effect of each module, different 3D backbone encoders and instance segmentation quality. Finally, we apply PAI3D to a proprietary autonomous driving dataset. The presented experimental results demonstrate that PAI3D improves 3D detection at far ranges and for small and thin objects.

  \section{Related Work}
  3D object detection has been thoroughly studied in recent years. Besides the large body of works that solely employ 3D point cloud, many researches emerge to adopt multi-modal sensors, primarily LiDAR and camera, to do 3D object detection (also called fusion-based methods \cite{MV3D,AVOD,3D-CVF,yang2018ipod,PointPainting,ding20201st,FusionPainting,mvp,park2021multi}), which utilize image semantics for improved 3D detection accuracy. Our PAI3D method belongs to this category of multi-modal fusion-based 3D object detection.

  \subsection{LiDAR Based 3D Object Detection}
  LiDAR-based 3D detection methods employ point cloud as the sole input, which can be roughly categorized into three categories of view-based, point-based, and voxel-based. View-based methods project points onto either Bird's-eye View (BEV) (PointPillar \cite{pointpillars}, PIXOR \cite{pixor}, Complex-YOLO \cite{complex-yolo}, Frustum-PointPillars \cite{Frustum-pointpillars}) or range view (RSN \cite{RSN}, RangeDet \cite{Rangedet}, RangeIoUDet \cite{RangeIoUDet}) to form a 2D feature map, 2D CNNs are then performed in this feature map to detect 3D objects. Point-based methods (PointRCNN \cite{shi2019pointrcnn}, STD \cite{yang2019std}, 3DSSD \cite{yang20203dssd}) directly predict 3D objects from the extracted point cloud features \cite{qi2017pointnet}. Such methods typically employ down sampling \cite{yang20203dssd} or segmentation \cite{shi2019pointrcnn,yang2019std} strategies to efficiently search for the objects in the point cloud. Voxel-based methods utilize 3D convolution (VoxelNet \cite{voxelnet}) or 3D sparse convolution \cite{spconv} (SECOND \cite{yan2018second}, SASSD \cite{sassd}, CenterPoint \cite{centerpoint}) to encode 3D voxelized features, and most succeeding efforts focus on exploring auxiliary information \cite{centerpoint,sassd} to enhance detections from voxelized data.

  \subsection{Multi-modal Fusion Based 3D Object Detection}
  Recently, more and more researches have emerged to combine LiDAR and camera sensing for 3D object detection, where rich visual semantics from images are fused with point cloud 3D geometry to improve performance. We group these works into two categories based on how fusion is performed: One-shot fusion methods \cite{MV3D,AVOD,3D-CVF} exploit implicit information embedded in image features and adopt an end-to-end deep NN to fuse two modalities and predict 3D bounding boxes. Sequential fusion methods \cite{yang2018ipod,PointPainting,ding20201st,FusionPainting,mvp,park2021multi} first predict semantic labels from images, and then augment LiDAR points with these labels by point-to-pixel projection, visual semantics are explicitly exploited in these methods and 3D detection are done in point cloud space once semantic labels are acquired at the first phase.

  Among one-shot fusion methods, MV3D \cite{MV3D} builds proposal-level fusion by generating proposals from LiDAR BEV followed by multi-view feature map fusion. AVOD \cite{AVOD} implements feature-level fusion where two identical but separate feature extractors are used to generate feature maps from both the LiDAR BEV map and the image followed by multi-modal feature fusion with a region proposal network to generate 3D proposals. Note that the misalignment crossing multiple views, i.e. LiDAR BEV and front view, and camera perspective view, can pose great challenges to MV3D and AVOD where the inconsistency in the fused features will degrade final detection performance. 3D-CVF \cite{3D-CVF} improves the cross-view misalignment issue by representing point cloud and images in the BEV feature maps so that the two modalities can be well aligned. Nevertheless, the lack of depth in images makes the conversion from camera perspective view to LiDAR BEV less accurate.

  Among sequential fusion methods, IPOD \cite{yang2018ipod} utilizes image semantic segmentation to label the foreground and background in LiDAR points, then it detects objects from only the foreground points. PointPainting \cite{PointPainting} further attaches categorical semantic labels to each LiDAR point, which improves mAP significantly. HorizonLiDAR3D \cite{ding20201st} adopts similar idea to PointPainting, but it employs object detection instead of image semantic segmentation to label the points encompassed by the detection bounding boxes. Inspired by PointPainting, PointAugmenting \cite{wang2021pointaugmenting} replaces the hard-categorical labels with softening features extracted from the backbone, leading to further improved detection performance. FusionPainting \cite{FusionPainting} applies semantic segmentation to both image and point cloud, and an attention mechanism is then introduced to fuse the two modalities. MVP \cite{mvp} utilizes image instance segmentation to generate virtual points to make object point cloud denser and hence easier to be detected. MTC-RCNN \cite{park2021multi} uses PointPainting in a cascaded manner that leverages 3D box proposals to improve 2D segmentation quality, which are then used to further refine the 3D boxes.

  Most existing fusion-based approaches concentrate on feature-level or pixel-to-point-level fusion from image to point cloud. The proposed PAI3D is a sequential fusion method that utilizes instance-level image semantics as 3D instance priors to guide 3D object detection, these 3D instance priors include object category label, the point-to-object membership and object position. Our ablation experiment results show that these instance priors are able to greatly boost 3D object detection performance.

  \section{Painting Adaptive Instance for Multi-modal 3D Object Detection}
  \label{pai3d}
  \begin{figure}[htbp]
    \centering
    \includegraphics[width=12.cm]{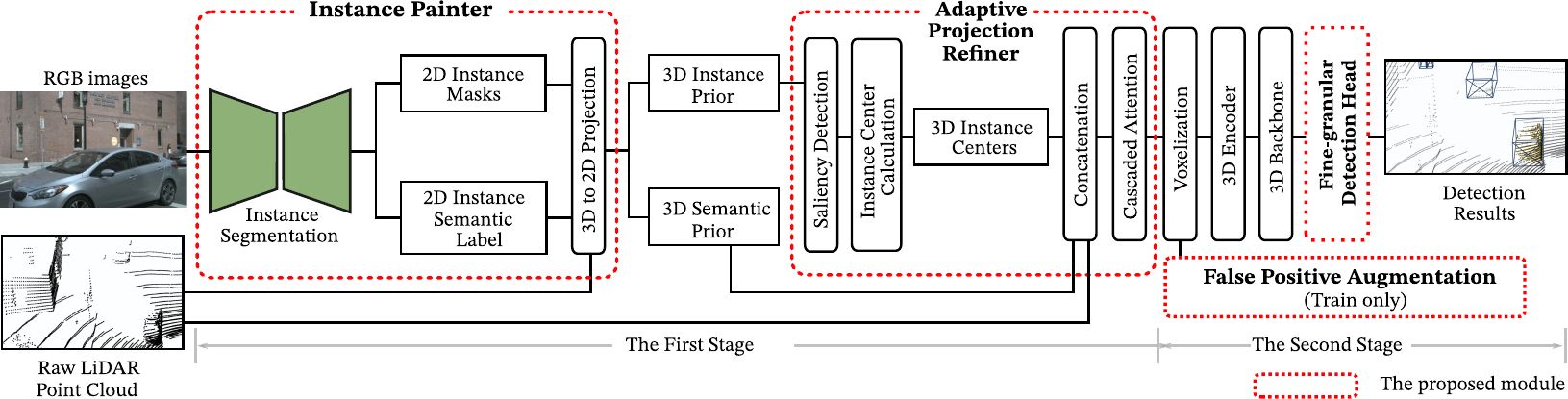}
    \caption{PAI3D architecture overview. PAI3D consists of three main components: Instance Painter extracts 3D instance and semantic priors from image instance segmentation result, Adaptive Projection Refiner corrects the erroneous semantic labels caused by projection mismatch and extract a center for a 3D instance, and Fine-granular Detection Head is capable of detecting objects with large-scale variations. False Positive Augmentation (FPA) serves as an additional training strategy to eliminate false positive proposals by online augmentation.}
    \label{fig:overall}
  \end{figure}
  In this section, we present the proposed PAI3D method, its architecture overview is shown in Fig. \ref{fig:overall}. Let $\ell(x, y, z, r)$ be a LiDAR point, where $x$, $y$, $z$ are the 3D coordinates and $r$ is the reflectance. Given a point cloud $L=\{\ell_i\}_{i=1}^N$, and $M$ views of images, $\{I_j\}_{j=1}^M$, that cover the 360-degree surrounding view, our goal is to utilize $\{I_j\}_{j=1}^M$ to obtain each point's semantic label $s$, the 3D instance $C$ such that $\ell_i \in C$, and the 3D instance center $\hat{C}=(C_x, C_y, C_z)$. We augment each raw point with semantic label and 3D instance center, that is $\ell_i$ is represented by $(x, y, z, r, s, C_x, C_y, C_z)$, and then we detect 3D objects from the set of augmented LiDAR points. To achieve this goal, we design Instance Painter and Adaptive Projection Refiner to obtain $(C, s)$ and $\hat{C}$ respectively. Moreover, we propose Fine-granular Detection Head with multiple feature maps of varying receptive fields to detect objects with varying scales.

  \subsection{Instance Painter}
\begin{figure}[h]
    \centering
    \includegraphics[width=12cm]{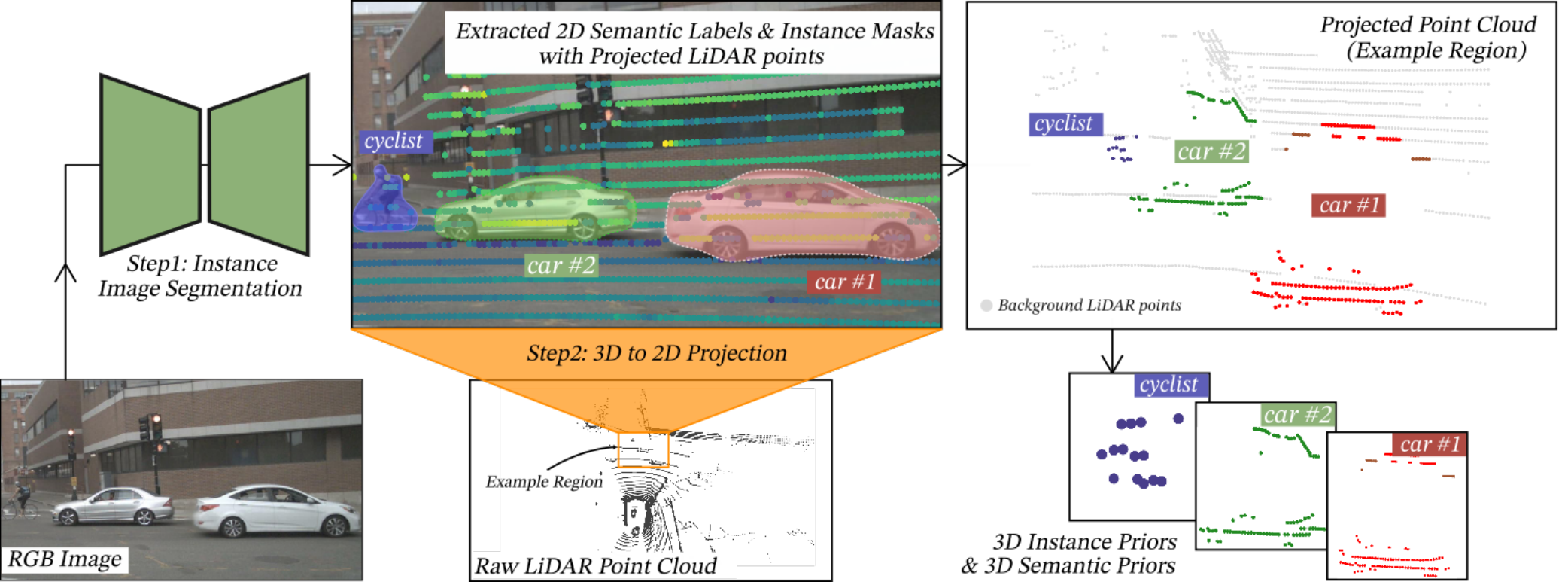}
    \caption{
      Instance Painter workflow. Best viewed in color. It aims to obtain the 3D object instances and their semantic labels to facilitate the 3D object detection task in the second stage. First, it uses image instance segmentation model to extract the 2D instance masks and semantic labels. Then it projects the point cloud onto the 2D instance masks to obtain 3D instance priors with semantic labels. However, the aggregated 3D instance may contain points with erroneous semantic labels, e.g. the points at the top of car $\#1$ actually are building points. This problem is solved by Adaptive Projection Refiner.
    }
    \label{fig:instance_painter}
  \end{figure}

  The goal of Instance Painter is to obtain the 3D object instances and their semantic labels (e.g. Car, Cyclist) as instance priors from point cloud utilizing image as ``assistive'' input (shown in Fig. \ref{fig:instance_painter}). There are three steps to achieve this goal as described in the following.

  First, image instance segmentation is performed to extract the 2D instance masks and semantic labels from images. Comparing with image semantic segmentation, instance segmentation can provide not only instance-aware semantic labels but also clear object instance delineation, allowing us to more accurately aggregate points belonging to the same 2D instance into one 3D instance in the subsequent instance aggregation step. In addition, the detailed visual texture, appearance, and color information in images can yield much higher instance segmentation performance than instance segmentation from point cloud.

  Second, Instance Painter performs a frustum-based 3D to 2D projection \cite{PointPainting,mvp} to project LiDAR points to images. The transformation from LiDAR coordinate to image coordinate is defined as $KT_{(cam \leftarrow ego)}T_{(ego^{t1} \leftarrow ego^{t2})}T_{(ego \leftarrow lidar)}$, where $T_{(ego \leftarrow lidar)}$, $T_{(ego^{t1} \leftarrow ego^{t2})}$ and $T_{(cam \leftarrow ego)}$ are the transformations from LiDAR frame to ego frame, from ego frame at time $t_2$ to that at $t_1$, and from ego frame to the target camera frame, respectively. $K$ denotes the target camera's intrinsic matrix.

  Finally, Instance Painter outputs a 3D instance prior by aggregating the LiDAR points which are projected into the same 2D object instance. Note that, two special cases need to be taken care of: first, the front and rear portion of an object can appear in two images of adjacent camera views due to object-at-image-border truncation, in this case we merge the two point clusters projected into the two images of the same object into one complete 3D instance; the second case is that, an object might be projected into two images of adjacent camera views if it occurs in the overlapped field of view of two adjacent cameras, and hence the 2D instance label and instance mask predicted from the two images might conflict with each other, in this case the instance label and instance mask having higher segmentation score is chosen for 3D instance aggregation.

  \subsection{Adaptive Projection Refiner}
\begin{figure}[t]
    \centering
    \includegraphics[width=12.cm]{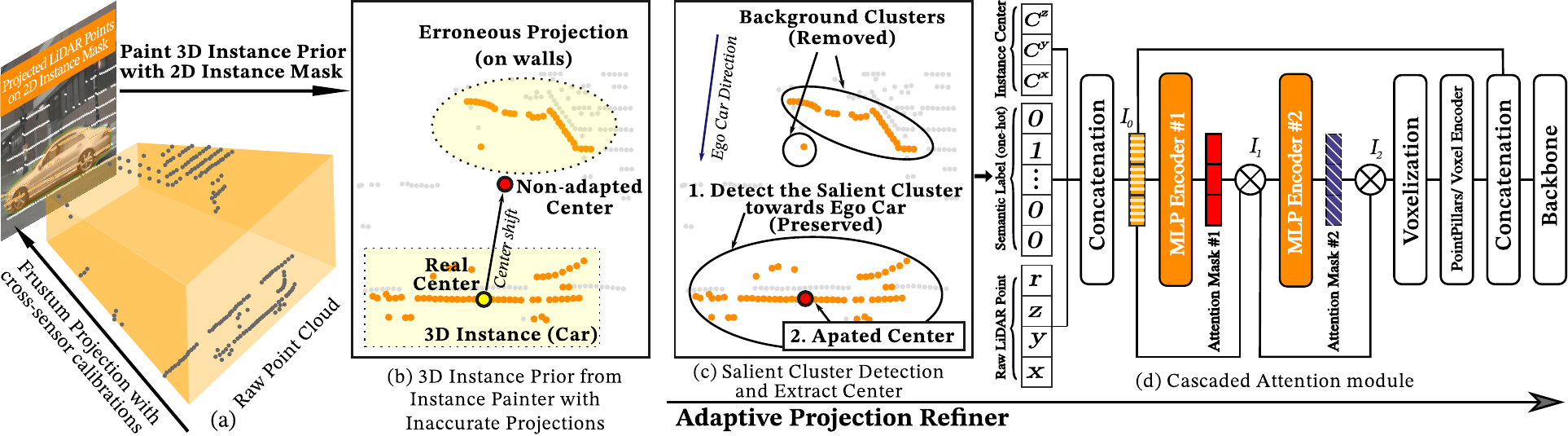}
    \caption{Adaptive Projection Refiner. 3D instance priors obtained from Instance Painter may not accurately capture object's true shape and scale due to many inevitable factors. Mean center usually deviates too much from the real position of the 3D instance center (a) and (b). Adaptive Projection Refiner refines the shape of the 3D instance by detecting the salient cluster closest to the ego car, computing cluster medoid as the 3D instance center (c), and employing a cascaded attention module to adaptively fuse the features from all modalities (d).
    }
    \label{fig:refiner}
  \end{figure}

  In a real-world autonomous car, many factors may inevitably lead to erroneous semantic labels in a 3D instance prior, these factors include imprecise 2D instance segmentation (especially at object boundary), object occlusions, point-to-pixel projection errors caused by inaccurate sensor calibration and/or multi-sensor synchronization offset. In addition, the projection path from point cloud to image is a frustum-shaped 3D subspace, implying that all the points in this subspace will be projected to the same instance mask (Fig. \ref{fig:refiner} (a)), which makes it prone to merge a 3D instance with points from a distant object (Fig. \ref{fig:refiner} (b)). Therefore, if a mean center is directly calculated from the projected points, this mean center will deviate too much from the true 3D instance center (Fig. \ref{fig:refiner} (b))).

  To solve these problems, we design Adaptive Projection Refiner to refine the 3D instance prior that may contain erroneous semantic labels and compute the medoid as the center vector to represent the 3D instance. A cascaded attention module is then proposed to adaptively fuse the raw LiDAR points, semantic labels, and instance centers.

  \textbf{Salient Cluster Detection}. The discovered 3D instances and the background LiDAR points with erroneous labels can manifest in arbitrary shapes and scales. We propose to utilize density-based spatial clustering, i.e., DBSCAN \cite{dbscan} or its efficient variants \cite{mr-dbscan,liu-landmark-fn-dbscan}, to refine the 3D instance shape by detecting the salient cluster towards the ego car and removing the background cluster. The salient cluster here denotes the cluster with the largest number of LiDAR points and closest to the ego car. Moreover, salient cluster medoid is computed to represent the 3D instance center since medoid is more robust to outliers compared with mean center (Fig. \ref{fig:refiner} (c)).

  \textbf{Cascaded Attention}. We design a cascaded channel attention mechanism to learn how to adaptively fuse the raw LiDAR points, semantic labels, and the 3D instance centers, which are fed to this module as channels. Specifically, each attention component uses a multilayer perceptron (MLP) to obtain a weight mask on all channels and then it produces features as weighted sum of the channels. These learned features are then encoded by a PointPillars \cite{pointpillars} or VoxelNet \cite{voxelnet} encoder. To preserve point-level features, we use skip connection to concatenate the voxelized features with the raw LiDAR point (Fig. \ref{fig:refiner} (d)).

  \textbf{Discussion}. The 3D instance center plays two important roles in 3D object detection. First it represents the membership of a point to a potential object, allowing the detector more easily distinguish the boundary of different objects. Second it encodes object position and provides significant clue to the location regression task, leading to increased position prediction accuracy in 3D object detection.

  Instance Painter and Adaptive Projection Refiner form the first stage of PAI3D, it produces tremendous guidance information to the 3D detection network in the second stage, including object semantic label, point-to-object membership, and object position, before 3D detection network even begins to detect 3D objects. Experimental results in Section \ref{sec:experiments} prove that the two modules in the first stage bring significant 3D detection improvement.

  \subsection{Additional Enhancements for 3D Object Detection}
  \begin{figure}[t]
    \centering
    \includegraphics[width=10.cm]{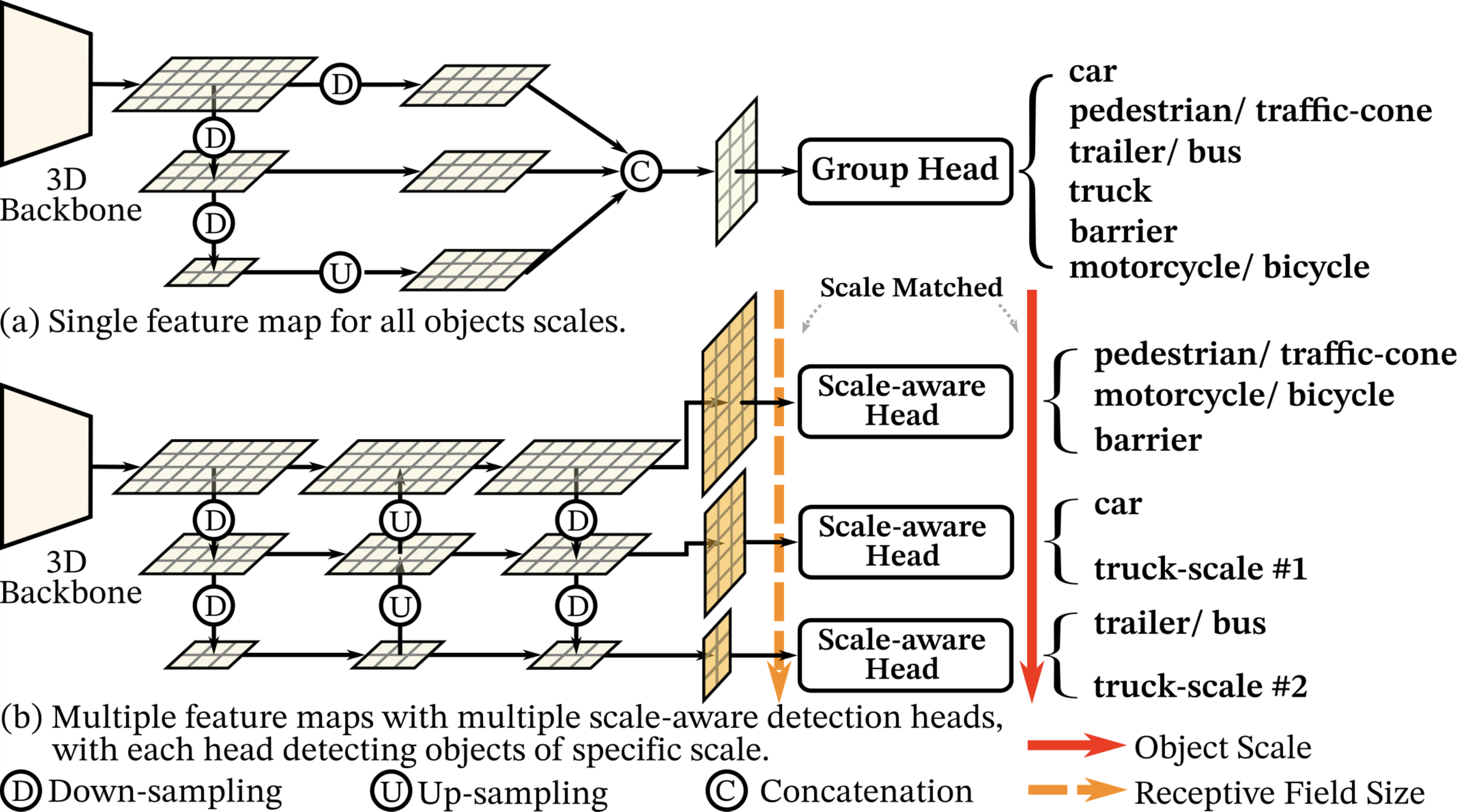}
    \caption{A popular detection head structure (a) uses a single aggregated feature map to detect objects of all scales. Our proposed Fine-granular Detection Head (b) picks up a feature map of appropriate receptive field from pyramid feature maps to detect objects with specific scale. The size of the receptive field of a feature map is positively correlated with the object scale.}
    \label{fig:fdh}
  \end{figure}
  \textbf{Fine-granular Detection Head}
  Detecting objects with various scales is crucial for a modern 3D detector. A learning strategy proposed recently, Group Head \cite{CBGS} divides object categories into serval groups based on object shapes or sizes, a detection head is then assigned to capture the intra-scale variations among objects for each group. However, common Group Head utilizes a single feature map as input, indicating the features it receives from the backbone are on a single-level scale. Most recently proposed detectors, e.g., CenterPoint \cite{centerpoint} and MVP \cite{mvp}, aggregate three scale levels into one feature map to alleviate this problem. However, they still use one feature map for all object scales (Fig. \ref{fig:fdh} (a)). A recent study (YOLOF \cite{yolof}) shows that passing multiple scaled feature maps to multiple detecting heads is essential in improving the detection performance.

  We propose Fine-granular Detection Head that detects specific-scaled objects with specific-scaled feature maps (Fig. \ref{fig:fdh} (b)). In details, an FPN-like \cite{fpn} (i.e. PANet \cite{panet}, BiFPN \cite{bifpn}) detection neck is used to generate a series of feature maps with different scales of receptive-fields, and then a Group Head is attached to each generated feature map so that each detection head becomes scale aware. Consequently, we can then dispatch different scaled objects of different categories into different scale-aware heads. Under this design, a scale-aware head naturally picks up a feature map with an appropriately scaled receptive field to match with the object scale. Likewise, this module also improves detection performance of the objects with large intra-scale (same category) variations by simultaneously learning the intra-scale variance of one category with multiple detection heads.

  \textbf{False Positive Augmentation}
  Typically, sparse LiDAR points are more prone to generate false positive detections, hence we propose False Positive Augmentation to address this issue. After training from scratch is complete, a fine-tuning is invoked, at when we build up an offline database to collect false positive proposals by running inference on the whole training dataset. Then we randomly sample false positive examples from the database and paste them into the raw point cloud, which is repeated for each successive fine-tuning step. The pasted false positive examples do not need extra annotations as they are recognized as background LiDAR points. Our ablation Experiments prove the effectiveness of this strategy in boosting detection performance.

  \section{Experiments}
  \label{sec:experiments}
  We evaluate PAI3D against existing 3D object detection methods on the nuScenes \cite{nuscenes} dataset and present a series of ablation results to offer insights into the effect of each module, different 3D encoders, and instance segmentation quality.

  \subsection{The nuScenes Dataset}
  The nuScenes dataset \cite{nuscenes} is one of the most well-known public large-scale datasets for autonomous driving. It is widely used by researchers to evaluate common computer vision tasks such as 3D object detection and tracking. The data was collected by a 32-beam LiDAR with a scanning frequency of 20Hz and six RGB cameras with a capture frequency of 12Hz. It provides 3D annotations for ten classes with 2Hz in a 360-degree field of view. The dataset contains 40K complete annotated frames, further split into 700 training scenes, 150 validation scenes, and 150 testing scenes. The official evaluation metrics includes mean Average Precision (mAP) \cite{mAP} and NuScenes Detection Score (NDS) \cite{nuscenes}. No external data are used for the trainings and evaluations on the nuScenes dataset.

  \subsection{Training and Evaluation Settings}
  We train and evaluate our method on four distributed NVIDIA P40 GPUs which are configured to run synchronized batch normalization with batch size 4 per GPU. Following \cite{FusionPainting}, we use HTCNet\footnote{We directly use the checkpoint (trained on nuImage \cite{nuscenes}) provided by MMDetection3D \cite{mmdet3d2020} without making any modifications.} \cite{htc} for both image instance segmentation in our PAI3D and 2D semantic segmentation in PointPainting \cite{PointPainting}.

  For the 3D object detection task, following \cite{nuscenes,centerpoint,FusionPainting,mvp} we stack ten LiDAR sweeps into the keyframe sample to obtain a denser point cloud. We evaluate PAI3D's performance with two different 3D encoders, VoxelNet \cite{voxelnet} and PointPillars \cite{pointpillars}. For VoxelNet, the detection range is set to  $[-54, 54]\times[-54, 54]\times[-5, 3]$ in meters for $X$, $Y$, $Z$ axes respectively, and the voxel size is $(0.075, 0.075, 0.075)$ in meters. For PointPillars, the detection range is set to $[-51.2, 51.2]\times[-51.2, 51.2]\times[-5, 3]$ in meters with voxel size being $(0.2, 0.2, 8)$ in meters.

  \subsection{nuScenes Test Set Results}

  \begin{table}[t]
    \begin{center}
    \caption{
  The 3D object detection results on the nuScenes {\it test set}. Object categories include Car, truck (Tru), Bus, trailer (Tra), construction vehicle (CV), pedestrian (Ped), motorcycle (MC), bicycle(Bic), traffic cone (TC), and barrier (Bar). ``L'' and ``C'' in Modality refer to ``LiDAR'' and ``Camera'' respectively. The proposed PAI3D outperforms all existing 3D detection methods, including LiDAR-camera fusion-based and LiDAR alone methods, with a large margin on all object categories (except traffic cone) and with both mAP and NDS metrics.}

    \label{table:nu_test}
    \resizebox{12.1cm}{!} {
    \begin{tabular}{l|c|cc|llllllllll}
    \hline\noalign{\smallskip}
    Method & Modality & \textbf{mAP}$\uparrow$ & \textbf{NDS}$\uparrow$ & Car & Tru & Bus & Tra & CV & Ped & MC & Bic & TC & Bar\\
    \noalign{\smallskip}
    \hline
    \noalign{\smallskip}
    PointPillars \cite{pointpillars} & L & 30.5 & 45.3 & 68.4 & 23.0 & 28.2 & 23.4 & 4.1 & 59.7 & 27.4 & 1.1 & 30.8 & 38.9\\
    WYSIWYG \cite{wysiwyg} & L & 35.0 & 41.9 & 79.1 & 30.4 & 46.6 & 40.1 & 7.1 & 65.0 & 18.2 & 0.1 & 28.8 & 34.7\\
    PointPainting \cite{PointPainting} & L \& C & 46.4 & 58.1 & 77.9 & 35.8 & 36.1 & 37.3 & 15.8 & 73.3 & 41.5 & 24.1 & 62.4 & 60.2\\
    CBGS \cite{CBGS} & L & 52.8 & 63.3 & 81.1 & 48.5 & 54.9 & 42.9 & 10.5 & 80.1 & 51.5 & 22.3 & 70.9 & 65.7\\
    CVCNET \cite{CVCNET} & L & 55.8 & 64.2 & 82.6 & 49.5 & 59.4 & 51.1 & 16.2 & 83.0 & 61.8 & 38.8 & 69.7 & 69.7\\
    CenterPoint \cite{centerpoint} & L & 58.0 & 65.5 & 84.6 & 51.0 & 60.2 & 53.2 & 17.5 & 83.4 & 53.7 & 28.7 & 76.7 & 70.9 \\
    HotSpotNet \cite{HotSpotNet} & L & 59.3 & 66.0 & 83.1 & 50.9 & 56.4 & 53.3 & 23.0 & 81.3 & 63.5 & 36.6 & 73.0 & 71.6\\

    MVP \cite{mvp} & L \& C & 66.4 & 70.5 & 86.8 & 58.5 & 67.4 & 57.3 & 26.1 & 89.1 & 70.0 & 49.3 & 85.0 & 74.8 \\
    CenterPointV2 \cite{centerpoint} & L \& C & 67.1 & 71.4 & 87.0 & 573. & 69.3 & 60.4 & 28.8 & 90.4 & 71.3 & 49.0 & \textbf{86.8} & 71.0 \\
    FusionPainting \cite{FusionPainting} & L \& C & 68.1 & 71.6 & 87.1 & 60.8 & 68.5 & 61.7 & 30.0 & 88.3 & 74.7 & 53.5 & 85.0 & 71.8 \\
    \hline
    \textbf{PAI3D (Ours)} & L \& C & \textbf{71.4} & \textbf{74.2} & \textbf{88.4} & \textbf{62.7} & \textbf{71.3} & \textbf{65.8} & \textbf{37.8} & \textbf{90.3} & \textbf{80.8} & \textbf{58.2} & 83.2 & \textbf{75.5} \\
    \hline

    \end{tabular}
    }
    \end{center}
  \end{table}

  Table \ref{table:nu_test} shows the results of PAI3D in comparison to the recent SOTA approaches on the nuScenes {\it test} set. Following \cite{centerpoint,FusionPainting}, we perform test time augmentation (TTA) operations including flip and rotation, and we apply NMS to merge the results from five models with different voxel sizes. As seen from Table \ref{table:nu_test}, PAI3D achieves 71.4 mAP and 74.2 NDS, outperforming all existing 3D detection methods, including LiDAR-camera fusion based and LiDAR alone methods, with a large margin on all object categories (except traffic cone) and with both mAP and NDS metrics. Note that latest methods such as CenterPointV2 (CenterPoint \cite{centerpoint} with PointPainting \cite{PointPainting}) and FusionPainting \cite{FusionPainting} are already very strong, nevertheless PAI3D outperforms these two strong baselines by 4.3 in mAP and 2.8 in NDS.

\begin{figure}[t]
    \centering
    \includegraphics[width=6cm]{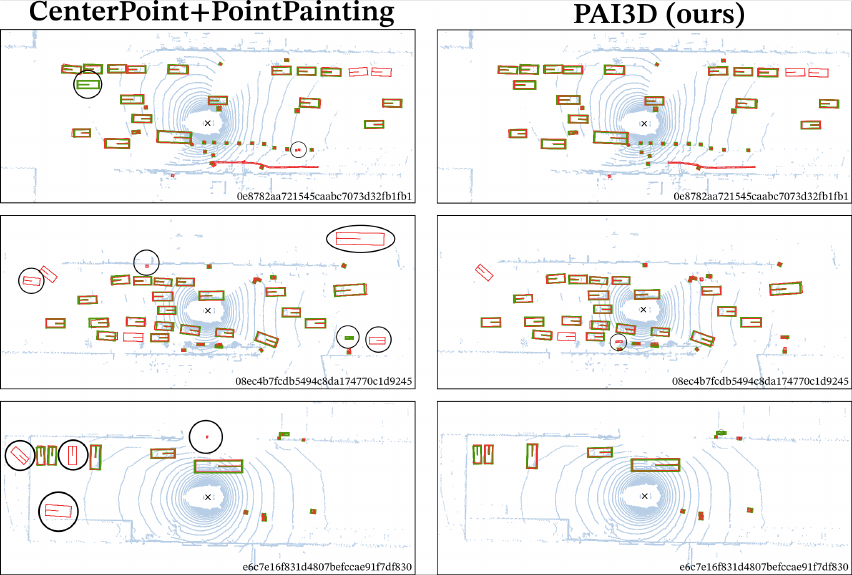}
    \caption{
      Qualitative results of PAI3D compared with CenterPoint (reproduced) + PointPainting applied to the nuScenes {\it validation} set. Predicted boxes are shown in red, ground truth boxes in green, and detection differences in black circles.
    }
    \label{fig:qual_results_nu}
  \end{figure}

  Qualitative results of PAI3D in comparison to strong SOTA CenterPoint \cite{centerpoint} (reproduced\footnote{Reproduced with the latest code from the CenterPoint's official GitHub repository: https://github.com/tianweiy/CenterPoint.}) + PointPainting (with HTCNet \cite{htc}) are presented in Fig.\ref{fig:qual_results_nu} (a). The results show that instance-level semantics fused with point geometry in PAI3D leads to fewer false positive detections and higher recall for the sparse objects

  \subsection{Ablation Study}
  In this section, we design a series of ablation studies using the nuScenes {\it validation} set to analyze the contribution of each module in the proposed method.

  \setlength{\tabcolsep}{4pt}
  \begin{table}[t]
    \begin{center}
    \caption{
      Ablation results for contribution of each module in PAI3D on the nuScenes {\it validation} set. FDH: Fine-granular Detection Head, IP: Instance Painter, SL: Semantic Label, Cen: Instance Center, APR: Adaptive Projection Refiner, FPA: False Positives Augmentation. The baseline is the reproduced CenterPoint with PointPillars encoder.
    }
    \label{table:ab_study_modules}
    \resizebox{12.1cm}{!} {
    \begin{tabular}{lc|cc|c|c|c|c}
    \hline\noalign{\smallskip}
    \multirow{2}{*}{Method} & \multirow{2}{*}{FDH} & \multicolumn{2}{|c|}{IP} & \multirow{2}{*}{APR} & \multirow{2}{*}{FPA}  & \multirow{2}{*}{\textbf{mAP}$\uparrow$}  & \multirow{2}{*}{\textbf{NDS}$\uparrow$}\\
  &  & SL & Cen &  &  &  & \\
    \noalign{\smallskip}
  \hline
    CenterPoint \cite{centerpoint} \space & & & & & & 50.3 & 60.2\\
    CenterPoint (reproduced baseline) & & & & & & 52.1 & 61.4\\
  \textbf{PAI3D (ours)} & \checkmark & & & & & 52.6 (0.5$\uparrow$) & 62.3 (0.9$\uparrow$)\\
  & \checkmark & \checkmark &  & & & 62.2 (9.6$\uparrow$)& 66.9 (4.6$\uparrow$)\\
  & \checkmark & \checkmark &  \checkmark & & & 64.1 (1.9$\uparrow$)& 67.7 (0.8$\uparrow$)\\
  & \checkmark & \checkmark & \checkmark & \checkmark & & 64.5 (0.4$\uparrow$)& 68.2 (0.5$\uparrow$)\\
  & \checkmark & \checkmark & \checkmark & \checkmark & \checkmark & 64.7 (0.2$\uparrow$)& 68.4 (0.2$\uparrow$)\\
  \hline
    \noalign{\smallskip}
    \end{tabular}
    }
    \end{center}
  \end{table}
  \setlength{\tabcolsep}{1.4pt}
  \noindent
  \textbf{Contribution of Each Module}. Incremental experiments are performed to analyze the effectiveness of each module by benchmarking against reproduced CenterPoint \cite{centerpoint} with PointPillars \cite{pointpillars} encoder as the baseline. As seen from Table \ref{table:ab_study_modules}, Fine-granular Detection Head boosts over the single Group Head \cite{CBGS} used in the baseline by 0.5 mAP and 0.9 NDS. Notably the highest performance boost comes from the Instance Painter (IP), which increases mAP by 11.5 and NDS by 5.4 over baseline with FDH, suggesting that instance-level image semantics are tremendously helpful. Within IP, semantic label brings 9.6 mAP and 4.6 NDS boost and using instance-center further improves performance by 1.9 mAP and 0.8 NDS. Adaptive Projection Refiner yields slight increase of 0.4 mAP and 0.5 NDS, indicating that our PAI3D, even without refining the 3D instances, is still quite robust to the erroneous semantic labels. False Positives Augmentation contributes an additional 0.2 mAP and 0.2 NDS. In total, our method outperforms the baseline by 12.6 mAP and 7.0 NDS, enabling PAI3D to be the new STOA method in fusion-based 3D object detection.

  \noindent
  \textbf{3D Encoders}. As shown in PAI3D architecture in Fig. \ref{fig:overall}, 3D encoders fuse the instance-level image semantics with point geometry features, and we are interested to see how sensitive PAI3D is to the choice of different 3D encoders. To that end, we compare PAI3D with strong fusion-based SOTA methods, including CenterPoint \cite{centerpoint} (reproduced) + PointPainting \cite{PointPainting} (HTCNet), CenterPoint + virtual-point \cite{mvp} and FusionPainting \cite{FusionPainting}, and we select two widely used 3D encoders, VoxelNet \cite{voxelnet} and PointPillars \cite{pointpillars}, as the 3D encoders for these methods. For the sake of fair comparison, we do not use any two-stage refinement of the detected 3D proposals or dynamic voxelizations. As seen from Table 3, PAI3D consistently surpasses all existing fusion based SOTA methods for both 3D encoders with a large margin, indicating that the strength of PAI3D mainly arise from non-3D-encoder modules such as Instance Painter and its performance is less influenced by the 3D encoders. Notice that PAI3D, even only using semantics from image, still outperforms FusionPainting (2D+3D Painting) \cite{FusionPainting}, which uses semantics from both image and point cloud segmentation, for the VoxelNet encoder.

  \setlength{\tabcolsep}{4pt}
  \begin{table}[t]
    \begin{center}
    \caption{Ablation results for 3D encoders on the nuScenes {\it validation} set. We implement PAI3D with two widely used 3D encoders, VoxNet and PointPillars, and benchmark its performances on nuScenes validation set with SOTA fusion-based methods. ``L'' and ``C'' in Modality refer to ``LiDAR'' and ``Camera'' respectively.}
    \label{table:ab_study_encoders}
    \resizebox{12.1cm}{!} {
    \begin{tabular}{l|c|c|cc}
    \hline\noalign{\smallskip}
    Method & Modality& 3D Encoders & \textbf{mAP}$\uparrow$  & \textbf{NDS}$\uparrow$\\
    \noalign{\smallskip}
    \hline
    \noalign{\smallskip}
    CenterPoint \cite{centerpoint} & L & VoxNet & 56.4 & 64.8\\
    CenterPoint (reproduced)  & L & VoxNet & 58.9 & 66.5\\
    CenterPoint (reproduced) + PointPainting(HTCNet)  & L \& C & VoxNet & 64.9 & 69.3\\
    CenterPoint + virtual-point \cite{mvp} & L \& C & VoxNet & 65.9 & 69.6\\
    FusionPainting(2D Painting) \cite{FusionPainting} & L \& C & VoxNet & 63.8 & 68.5\\
    FusionPainting(2D+3D Painting) \cite{FusionPainting} & L \& C & VoxNet & 66.5 & 70.7\\
    \textbf{PAI3D (ours)} & L \& C & VoxNet & \textbf{67.6}	& \textbf{71.1}\\
    \hline
    \hline
    CenterPoint \cite{centerpoint} & L & PointPillars & 50.3 & 60.2\\
    CenterPoint (reproduced) & L & PointPillars & 52.1 & 61.4\\
    CenterPoint (reproduced) + PointPainting(HTCNet) & L \& C & PointPillars & 61.8 & 66.2\\
    \textbf{PAI3D (ours)} & L \& C & PointPillars & \textbf{64.7} &	\textbf{68.4}\\
    \hline
    \end{tabular}
    }
    \end{center}
  \end{table}
  \setlength{\tabcolsep}{1.4pt}

  \begin{figure}[b]
    \centering
    \includegraphics[height=3.3cm]{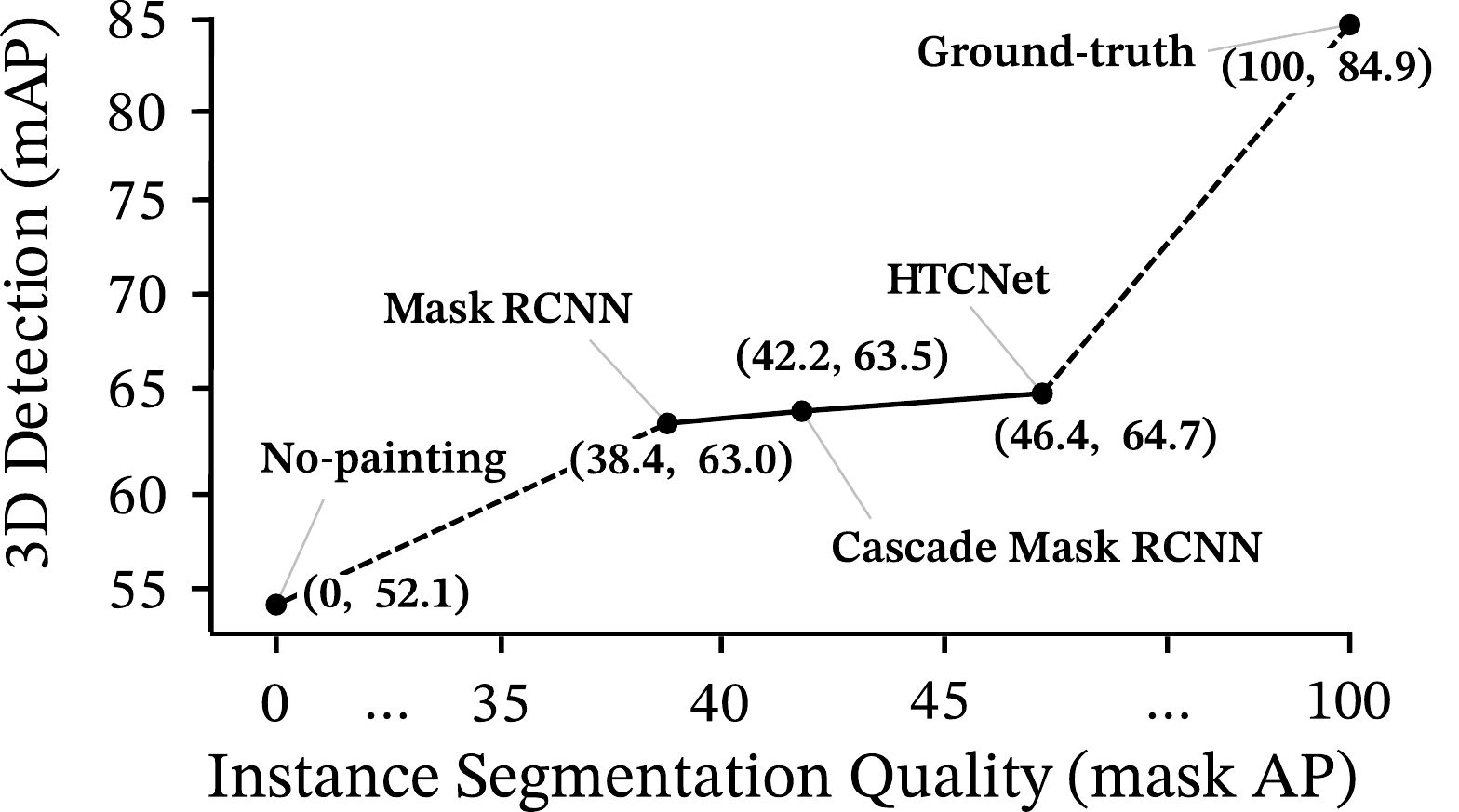}
    \caption{Ablation results for instance segmentation quality. Instance segmentation quality and 3D detection performance are positively correlated, meaning that better 2D instance segmentation leads to better 3D object detection performance.
    }
    \label{fig:dep_2d_model}
  \end{figure}

  \noindent
  \textbf{Instance Segmentation Quality}. \cite{PointPainting} shows that image segmentation quality can dramatically impact the final 3D object detection performance. To examine this issue, we conduct ablation experiments to plot the correlation curve between instance segmentation quality (measured by mask AP) and 3D detection performance (measured by mAP). We select several instance segmentation models with increasing quality, ranging from Mask R-CNN \cite{maskrcnn} to Cascade Mask R-CNN \cite{cascademaskrcnn} to HTCNet \cite{htc} from MMDetection3D \cite{mmdet3d2020}, and implement them into PAI3D, we then add the lower bound of no semantic painting and the upper bound of painting with ground truth label. All the instance segmentation models are trained on the nuImage \cite{nuscenes} dataset, and all these 3D detection methods are trained with nuScenes training dataset. Finally, we obtain the 3D object detection mAP of these methods on the nuScenes {\it validation} set. As shown in Fig. \ref{fig:dep_2d_model}, instance segmentation quality and 3D detection performance are positively correlated, meaning that better 2D instance segmentation leads to better 3D object detection performance. In addition, we observe that model-based painting yields much worse 3D detection performance than ground-truth-based painting, we point out that this performance gap can be narrowed by improving instance segmentation quality and reducing point-to-pixel projection errors. We believe more in-depth research along this direction is worthwhile to pursue in our future work.

  \section{Conclusions}
  This paper presents PAI3D, a simple and effective sequential instance-level fusion method for 3D object detection, which realizes instance-level fusion between 2D image semantics and 3D point cloud geometry features. PAI3D outperforms all the SOTA 3D detection methods, including LiDAR-camera fusion based and LiDAR alone methods, with a large margin on almost all object categories on the nuScenes dataset. Our thorough ablation experiments show that instance-level image semantics contributes the most to the performance gain, and PAI3D works well with any good-quality instance segmentation models and any modern point cloud 3D encoder networks. Qualitative results on a real-world autonomous driving dataset demonstrate that PAI3D yields fewer false positive detections and higher recall for the distant sparse objects and hard to see objects.

\clearpage
%
%

\end{document}